\documentclass[12pt]{article}

\usepackage{amsbsy,amsmath,amsthm,amssymb}
\usepackage{graphics}
\usepackage[demo]{graphicx}
\usepackage{setspace}
\usepackage{natbib}
\usepackage{fullpage}
\usepackage{url}

\theoremstyle{plain}

\theoremstyle{definition}
\newtheorem{algorithm}{Algorithm}
\theoremstyle{remark}

\DeclareMathOperator{\tr}{tr}

\DeclareMathOperator{\Sp}{\mathbb{S}}

\DeclareMathOperator{\Ds}{\mathcal{D}}

\DeclareMathOperator{\Us}{\mathcal{U}}
\DeclareMathOperator{\Cs}{\mathcal{C}}
\DeclareMathOperator{\Ls}{\mathcal{L}}

\newcommand{\argmin}{\qopname\relax m{arg\,min}}

\newcommand{\fatdot}{\,\cdot\,}

\makeatletter
\renewcommand{\section}{\@startsection {section}{1}{\z@}%
    {-3.5ex \@plus -1ex \@minus -.2ex}%
    {2.3ex \@plus.2ex}%
    {\centering\normalfont\Large\bfseries\uppercase}}
\makeatother

\begin{document}

\begin{center}
\LARGE \bf Ridge Fusion in Statistical Learning
\end{center}
\begin{center}
Bradley S. \textsc{Price}, Charles J. \textsc{Geyer},
and Adam J. \textsc{Rothman}
\end{center}

\newcommand{\foo}{\thefootnote}
\renewcommand{\thefootnote}{\relax}
\footnotetext{Bradley S. Price, School of Statistics, University of Minnesota
    (E-Mail: \emph{price412@umn.edu}).
    Charles J. Geyer, School of Statistics, University of Minnesota
    (E-mail: \emph{geyer@umn.edu}).
    Adam J. Rothman, School of Statistics, University of Minnesota
    (E-mail: \emph{arothman@umn.edu}).}
\renewcommand{\thefootnote}{\fred}

\begin{abstract}
We propose a penalized likelihood method to jointly estimate multiple precision
matrices for use in quadratic discriminant analysis and model based clustering. A ridge penalty
and a ridge fusion penalty are used to introduce shrinkage and promote similarity between precision matrix estimates.  
Block-wise coordinate descent is used for optimization,
and validation likelihood is used for tuning parameter selection.
Our method is applied in quadratic discriminant analysis
and semi-supervised model based clustering.

\par
\medskip
\noindent \textbf{Key Words:} Joint inverse covariance matrix estimation; Discriminant analysis; Model based clustering; Semi-supervised learning.
% NO REFERENCES in abstract (instructions for authors)
% Keywords are NOT contained in the title (instructions for authors)
% Look at a copy of the journal for formattting details not in instructions
%    for authors
\end{abstract}

\section{Introduction}
\label{sec:Intro}
Classification by quadratic discriminant analysis (QDA) requires the estimation
 of multiple inverse covariance matrices. 
In this model, the data $(x_1,y_1),\ldots, (x_n,y_n)$ are assumed to be a realization of $n$ independent copies of the random pair $(X,Y)$, where 
$Y$ is supported on $\Cs=\{1,\ldots, C\}$ and 
$(X \vert Y=c) \sim N_p(\mu_{0c},\Theta_{0c}^{-1})$ for each $c\in\Cs$.  Let $n_c=\sum_{i=1}^n 1(y_i=c)$ be the sample size
for the $c$th class, 
let $\bar x_c= n_{c}^{-1} \sum_{i=1}^n x_i 1(y_i=c)$ be the observed sample mean for the $c$th class, and let
$$
S_c=\frac{1}{n_c}\sum_{i=1}^n (x_i-\bar{x}_c)(x_i-\bar{x}_c)^T 1(y_i=c), \qquad c\in\Cs,
$$
be the observed sample covariance matrix for the $c$th class.  
Simply inverting $S_c$ to estimate $\Theta_{0c}$ is problematic when $n_c$ is small and impossible when $p \geq n_c$.
\citet{Pour11} reviews several regularized covariance and inverse covariance estimators that could be used to estimate 
the $\Theta_{0c}$'s, but this would not exploit similarities between them.
Similarity between 
the $\Theta_{0c}^{-1}$'s and low condition
numbers for each $\Theta_{0c}$ are exploited in regularized discriminant analysis (RDA) \citep{RegDiscrim}, which 
estimates $\Theta_{0c}$ by inverting a linear combination of $S_c$, the identity matrix, and the observed pooled sample covariance matrix.

Minus 2 times the profile log-likelihood function,
profiling over mean and class probability parameters, is
% Charlie: don't think "up to constants" was needed.  Additive terms
% not containing the parameter can be dropped from a log likelihood
% and it is still a log likelihood (in my terminology)
\begin{equation} \label{eq:gee}
g(\Theta_1, \ldots, \Theta_C) = \sum_{c\in \Cs} n_c\{\tr(S_c\Theta_c)-\log\det(\Theta_c)\},
\end{equation}
where $\tr$ and $\det$ are the trace and determinant operators.
In a more general setting, 
\citet{JointEstimationMGM} and \citet{JGL} proposed estimates of $\Theta_{01}, \ldots, \Theta_{0C}$ by minimizing \eqref{eq:gee} 
plus penalties that promote entry equivalence across the estimates of $\Theta_{01}, \ldots, \Theta_{0C}$ and zero entries
within estimates of the $\Theta_{0c}$'s.  The penalty of \citet{JointEstimationMGM} promoted zero entry equivalence
across the inverse covariance estimates and the penalty of \citet{JGL}, called Fused Graphical Lasso (FGL), promoted zero and non-zero 
entry equivalence across the inverse covariance estimates.
This sparse regularization is aimed at estimating multiple Gaussian graphical models, but
is another natural regularization for QDA.

We propose estimates of $\Theta_{01}, \ldots, \Theta_{0C}$ that minimize $g$ 
plus ridge penalties to promote entry-wise similarity between the estimates of the inverse covariance matrices and entry shrinkage  
for each inverse covariance estimate,
which is yet another natural regularization for QDA.
Our simulations and data examples illustrate cases where our estimators 
perform competitively in QDA.  We also apply our method and FGL to model-based clustering.

Let $\lvert A \rvert_q$ denote the $q$-norm of the vector formed from all
the entries of the matrix $A$. Let $\Sp^p$ denote the set of symmetric $p \times p$ matrices,
and let $\Sp^p_+$ the set of symmetric $p \times p$ positive definite matrices.

Computing our estimates relies on evaluating the function $Q(\fatdot,\lambda) : \Sp^p \rightarrow \Sp^{p}$ defined by
\begin{equation} \label{ridge-update}
Q(S, \lambda) = \argmin_{\Theta\in\Sp_{+}^p}\bigl\{ \tr(\Theta S) - \log\det(\Theta) +  \lambda|\Theta|_{2}^2/2\bigr\}.
\end{equation}
\citet{Scout} used the optimization in \eqref{ridge-update} in the context of covariance-regularized regression, where $S$ is
an observed sample covariance matrix and $\lambda$ is a non-negative tuning parameter.
For $\lambda > 0$, they derived the closed-form solution
$$
Q(S, \lambda) = \frac{1}{2\lambda}V\{-D +(D^{2}  + 4\lambda I)^{1/2} \}V^T,
$$
where $S = VDV^T$ with $V$ orthogonal and $D$ diagonal.  Iterative algorithms that evaluate $Q(\fatdot,\lambda)$
include the Fused Graphical Lasso (FGL) algorithm of \citet{JGL} and an iterative algorithm developed by
\citet{RothmanForzani13} that solves a modified version of \eqref{ridge-update} in which
the term $\lambda|\Theta|_{2}^2/2$ is replaced by $\lambda \sum_{i,j}m_{ij}\theta_{ij}^2/2$, where the $m_{ij}$'s are user-specified non-negative 
penalty weights.

\section{Joint Estimation with Ridge Fusion}
\label{SupMethod}
\subsection{Method}
We propose the penalized likelihood inverse covariance estimates
\begin{equation}
\label{JOICERidge}
(\widehat\Theta_1, \ldots, \widehat\Theta_C)
= \argmin_{\Theta_c \in \Sp^p_+, c \in \Cs} \left\{ g(\Theta_1, \ldots, \Theta_C) +  \frac{\lambda_1}{2}\sum_{c\in\Cs} |\Theta_c |_{2}^2
 + \frac{\lambda_2}{4} \sum_{(c,m)\in\Cs\times \Cs} | \Theta_c -\Theta_{m}|_{2}^2\right\},
\end{equation}
where $\lambda_1$ and $\lambda_2$ are non-negative tuning parameters.  
The term multiplied by $\lambda_1$ is called the \emph{ridge penalty}, and
the term multiplied by $\lambda_2$ is called the \emph{ridge fusion penalty}.
The former shrinks the elements of each $\widehat\Theta_c$ toward 
zero, and the latter promotes
entry-wise similarity between $\widehat\Theta_1, \ldots, \widehat\Theta_C$.  
Although these estimates are not invariant to scaling of the variables,  invariance is easily achieved by
standardizing the variables and then rescaling appropriately.  The objective function in \eqref{JOICERidge}
is strictly convex, and, if $\lambda_1 > 0$, then the global minimizer exists
and is unique.
% We present an algorithm to solve \eqref{JOICERidge}
% in Section \ref{algsection} and continue by discussing some special cases.

If $\lambda_2=0$, 
then \eqref{JOICERidge} decouples into $C$ separate 
ridge penalized likelihood problems, which have solutions
$\widehat\Theta_c = Q(S_c, n_{c}^{-1}\lambda_1)$ for $c\in\Cs$.

As $\lambda_2$ goes to infinity, $(\widehat\Theta_1, \ldots, \widehat\Theta_C)$ 
converges to
$(\widehat\Theta_{1}^{\bullet}, \ldots, \widehat\Theta_{C}^{\bullet})$
defined to be the solution to \eqref{JOICERidge}
subject to the constraint  $\Theta_1=\cdots=\Theta_C$, which is
\begin{equation}
\label{JOICERidge-edge}
\widehat\Theta_{1}^{\bullet} = \cdots = \widehat\Theta_{C}^{\bullet} 
= \argmin_{\Theta \in \Sp^p_+} \left\{ g(\Theta, \ldots, \Theta) +  \frac{\lambda_1}{2} C |\Theta|_{2}^2\right\} 
= Q\left(\frac{1}{n}\sum_{c\in\Cs} n_c S_c;  \frac{\lambda_1 C}{n} \right).
\end{equation}
This ``edge case'' is important both for computational efficiency ---
solving \eqref{JOICERidge} is computationally unstable when either $\lambda_1$
or $\lambda_2$
is very large due to the limited precision of computer arithmetic ---
and because it is itself a parsimonious model appropriate for some data.

\subsection{Algorithm} \label{algsection}
% having an \eqref in a heading is bizarre, also unnecessary
% if the first subsection heading can be just "method" this can be just "algorithm
We solve \eqref{JOICERidge} using block-wise coordinate descent.
The objective function in \eqref{JOICERidge} is
\begin{equation} \label{eq:objfun}
   f(\Theta_1, \ldots, \Theta_C)
   =
   g(\Theta_1, \ldots, \Theta_C)
   +
   \frac{\lambda_1}{2}\sum_{c\in\Cs} |\Theta_c |_{2}^2
   +
   \frac{\lambda_2}{4} \sum_{(c,m)\in\Cs\times \Cs} | \Theta_c -\Theta_{m}|_{2}^2
\end{equation}
with $g$ defined by \eqref{eq:gee}.
The blockwise coordinate descent step minimizes this with respect to one
$\Theta_c$, leaving the rest fixed.  This step has a closed-form expression.
Differentiating \eqref{eq:objfun} with respect to $\Theta_c$ and setting
the result equal to zero gives
$$
  n_c ( S_c - \Theta_c^{-1} ) + \lambda_1 \Theta_c
  + \lambda_2  \sum_{m\in\Cs\setminus \{c\}} (\Theta_c - \Theta_m)
  =
  0
$$
and dividing through by $n_c$ gives
\begin{equation} \label{zerograd}
  \widetilde{S}_c - \Theta_c^{-1} + \tilde{\lambda}_{c}  \Theta_c = 0,
\end{equation}
where
\begin{subequations}
\begin{align}
   \widetilde{S}_c
   & =
   S_c - \frac{\lambda_2}{n_{c}} \sum_{m\in \Cs\setminus \{c\}}  \Theta_m
   \label{eq:s-twiddle}
   \\
   \tilde\lambda_{c} & = \frac{\lambda_1 + \lambda_2(C-1)}{n_{c}}
   \label{eq:lambda-twiddle}
\end{align}
\end{subequations}
and, since the left-hand side of \eqref{zerograd} is the same as the gradient
of the objective function of \eqref{ridge-update} with $S$ replaced by
$\tilde{S}_c$ and  $\lambda$ replaced by $\tilde{\lambda}_c$,
the solution to \eqref{zerograd},
considered as a function of $\Theta_c$ only,
is $Q(\widetilde{S}_c; \tilde\lambda_c)$.

\begin{algorithm} \label{thealg}
Initialize a convergence tolerance $\varepsilon$
and $\Theta_1, \ldots, \Theta_C$.
\begin{tabbing}
Compute $\tilde{\lambda}_1, \ldots, \tilde{\lambda}_C$ using \eqref{eq:lambda-twiddle}.
\\
\textbf{repeat}
\\
\qquad \= \textbf{for} $c \in \Cs$ 
\\
\> \qquad \= Compute $\widetilde{S}_c$ using \eqref{eq:s-twiddle}.
\\
\>\>Set $\Theta_{c}^{\text{old}}=\Theta_c$
\\
\> \> Set $\Theta_c := Q(\widetilde{S}_c; \tilde{\lambda}_c)$.
\\
\> \textbf{end for}
\\
\textbf{until}
\\
\> $\sum_{c\in \Cs} |\Theta_{c}^{\text{old}} - \Theta_{c}|_{1} < \varepsilon  \sum_{c\in \Cs} |(S_{c}\circ I)^{-1}|_1$
\\
\textbf{end repeat}
\end{tabbing}
\end{algorithm}

The computational complexity of the blockwise descent algorithm is $O(Cp^3)$.
% no mathcal for "big O".  Look at a math book.
The initial iterate for Algorithm \ref{thealg} could be selected depending on the size of $\lambda_2$:
when $\lambda_2$ is large,  one could initialize at the edge-case estimates 
defined in \eqref{JOICERidge-edge}; and when $\lambda_2$ is small, one could initialize at 
the solution to \eqref{JOICERidge} when $\lambda_2=0$. 

\subsection{Tuning Parameter Selection}
\label{secTune}
Tuning parameter selection for \eqref{JOICERidge} is done using
a validation likelihood.  This is a generalization of its use in the single 
precision matrix estimation problem \citep{huang06, RothmanForzani13}.
Randomly split the data into $K$ subsets, dividing each of the $C$ classes 
as evenly as possible.
Let the subscript $(v)$ index objects defined for the $v$th subset of the data,
and $(-v)$ index those defined for the data with the $v$th subset removed.
The validation likelihood score is
\begin{equation}
\label{SupVL}
    V(\lambda_1,\lambda_2)
    =
    \sum_{v=1}^K\sum_{c\in \Cs} n_{c(v)} \bigl\{
    \tr(S_{c(v)}\widehat{\Theta}_{c(-v)})-\log\det(\widehat{\Theta}_{c(-v)})
    \bigr\},
\end{equation}
noting that $\widehat{\Theta}_{c(-v)}$ depends on
$\lambda_1$ and $\lambda_2$ even though the notation does not indicate this.
Our selected tuning parameters are $\hat{\lambda}_1$ and $\hat{\lambda}_2$ defined as the
values of the tuning parameters that minimize \eqref{SupVL} over the
set of their allowed values.

\section{Differences between ridge fusion and RDA} \label{ridgevsrda}
To gain some understanding of the difference between our ridge fusion method 
and RDA, we consider the special case where there is no fusion.  
For RDA, this means that the coefficient multiplying the pooled sample covariance matrix
is 0, so its $c$th covariance matrix estimate is
$(1-\beta)S_c+ \beta\bar{d} I$,
 where $\beta \in [0,1]$ is a tuning parameter and $\bar{d}$ is the arithmetic mean of the eigenvalues of $S_c$. Our ridge fusion method without fusion is defined by
 \eqref{JOICERidge}, with $\lambda_2=0$.  Decompose $S_c=VDV^T$ with $V$ orthogonal and $D$ diagonal.
The $c$th covariance estimate for ridge fusion without fusion is
$$
V\left\{0.5 D  + 0.5\left( D^{2}  + 4 \lambda_1 n_{c}^{-1}  I \right)^{1/2} \right\}V^T,
$$
and the $c$th covariance estimate for RDA without fusion is
$V\left\{ (1-\beta)D+\beta\bar{d}I\right\}V^T$.  
Both estimates have the same eigenvectors as $S_c$, but
their eigenvalues are different.  RDA shrinks or inflates the eigenvalues of 
$S_c$ linearly toward their average $\bar d$. Ridge fusion inflates the eigenvalues of $S_c$
nonlinearly, where the smaller eigenvalues of $S_c$ are inflated more
than the larger eigenvalues.

\section{Extension to Semi-Supervised Model Based Clustering}
\label{sec:SSMBC}
\subsection{Introduction}
Just as in classification using QDA, semi-supervised model based clustering with Gaussian mixture models requires estimates for multiple inverse covariance matrices.
In the semi-supervised model, let $\Ls$ and $\Us$ be disjoint sets
of cardinality $n_L$ and $n_U$, respectively.
The data are random pairs $(X_i, Y_i)$, where for $i \in \Ls$
both $X_i$ and $Y_i$ are observed but for $i \in \Us$ only $X_i$ is observed
($Y_i$ is latent). We denote $\Ds$ as the observed data. Otherwise, the setup is as in Section~\ref{SupMethod}.
 
Let the conditional probability density function of $X_i$ given $Y_i = c$,
which, as in section~2, we assume is Gaussian, be denoted by
$\phi(\fatdot;\mu_c,\Theta_c)$, where $\mu_{c}$ is the mean vector
and $\Theta_c$ is the inverse covariance matrix.
% The marginal density for $X$ is given by
% \begin{equation*}
% f_{X}(x)=\sum_{c \in \Cs} \phi_c(x)\pi_c,
% \end{equation*}
% not referred to
Let $\pi_c$ denote the probability of $Y_i = c$.
And let
$\Psi=\{\Theta_1,\ldots,\Theta_c,\mu_1,\ldots,\mu_c,\pi_1,\ldots,\pi_c\}$
denote all the parameters.

The log-likelihood for the observed data $\Ds$ with parameters $\Psi$ is
\begin{equation}
\label{Dloglike}
l(\Psi)=\sum_{i \in \Ls} \log\{\pi_{y_i}\phi(x_i;\mu_{y_i},\Theta_{y_i})\}+\sum_{i \in \Us}\log\left\{\sum_{c \in \Cs} \pi_c\phi(x_i;\mu_c,\Theta_c)\right\},
\end{equation}
and the complete data log-likelihood (treating the unobserved data as if
it were observed) is
\begin{equation}
\label{SemiMBC}
h(\Psi)=\sum_{i \in \Ls \cup \Us}\log\{\pi_{y_i}\phi(x_i;\mu_{y_i},\Theta_{y_i})\}.
\end{equation} 
Methods proposed in \cite{L1MBC}, \citet{DiagCluster}, and \citet{UnconMBC} seek to estimate the parameters of \eqref{Dloglike} using a penalized EM algorithm with assumptions of a specific structure or sparsity on both means and inverse covariances.
We propose to estimate these parameters by maximizing 
\eqref{Dloglike} penalized by ridge or $l_1$ penalties to create the same
kind of shrinkage discussed in sections~\ref{sec:Intro}
% using methods presented in \citet{JGL} ??? really need to cite them again ???
and~\ref{SupMethod}.
 We also will address tuning parameter selection by introducing a validation likelihood that uses 
 the unlabeled data.

\subsection{Joint Estimation In Semi-Supervised Model Based Clustering}

% No our estimates maximize penalized (missing data) likelihood, not penalized
% complete data likelihood.
% We propose the parameter estimates,$\widehat{\Psi}$, that maximize the penalized complete data log likelihood defined as

The penalized log likelihood is
\begin{equation}
\label{JOICEmbc}
 l(\Psi)
-\frac{\lambda_1}{j}\sum_{c\in\Cs} |\Theta_c |_{j}^j
 - \frac{\lambda_2}{j^2} \sum_{(c,m)\in\Cs\times \Cs} | \Theta_c -\Theta_{m}|_{j}^j,
\end{equation}
for $j\in\{1,2\}$. When $j=1$ \eqref{JOICEmbc} uses the
Fused Graphical Lasso (FGL) penalty \citep{JGL},
and when $j=2$ \eqref{JOICEmbc} uses the
the ridge fusion method penalty of section~\ref{SupMethod}.
Here we are introducing these penalties
to semi-supervised model based clustering.

We use the penalized analog of the EM Algorithm to
find maximum penalized likelihood estimates of \eqref{JOICEmbc} \citep{DLR,EMAlg,PEMcite} .
Let $\widehat{\Psi}$ denote the current iterate of the parameter estimates.
Then the E-Step of the algorithm calculates
\begin{equation}
\label{EqnQ}
\begin{split}
Q_{\widehat{\Psi}}(\Psi)=&E_{\widehat{\Psi}}\left(h(\Psi)-\frac{\lambda_1}{j}\sum_{c\in\Cs} |\Theta_c |_{j}^j
 - \frac{\lambda_2}{j^2} \sum_{(c,m)\in\Cs\times \Cs} | \Theta_c -\Theta_{m}|_{j}^j\, \middle|\Ds\right),\\
=&\sum_{i \in \Ls}\log\{\pi_{y_i}\phi(x_i;\mu_{y_i},\Theta_{y_i})\}+
\sum_{i \in \Us}\sum_{c\in\Cs}\alpha_{ic}\log\{\pi_{c}\phi(x_i; \mu_c,\Theta_c)\}\\
&-\frac{\lambda_1}{j}\sum_{c\in\Cs} |\Theta_c |_{j}^j
 - \frac{\lambda_2}{j^2} \sum_{(c,m)\in\Cs\times \Cs} | \Theta_c -\Theta_{m}|_{j}^j,
\end{split}
\end{equation}
where
\begin{equation}
\alpha_{ic}=\frac{\phi(x_i;\hat{\mu}_c,\hat{\Theta}_c)\hat{\pi}_c}{\sum_{m\in \Cs}\phi(x_i;\hat{\mu}_m,\hat{\Theta}_m)\hat{\pi}_m},
 \qquad \text{$i \in \Us$ and $c \in \Cs$}.
\end{equation}

The M-Step of the algorithm calculates $\widehat{\Psi}$ that maximizes \eqref{EqnQ} with respect to $\Psi$.  
% The estimators for the parameters of $\widehat{\Psi}$ are
Define
\begin{align}
  \tilde{n}_c & = \sum_{i\in \Ls} 1(y_i=c)+\sum_{i \in \Us} \alpha_{ic}
   \nonumber
   \\
   \tilde{\pi}_c & = \frac{\tilde{n}_c}{n_L+n_U},
   \label{eq:pi-twiddle}
   \\
 \tilde{\mu}_c & =\frac{\sum_{i\in \Ls}x_{i}1(y_i=c)+
 \sum_{i \in \Us}\alpha_{ic}x_i}{\tilde{n}_c}
   \label{eq:mu-twiddle}
   \\
   \nonumber
\tilde{S}_c^{(L)}=&\frac{\sum_{i\in \Ls}1(y_i=c)(x_i-\tilde{\mu}_c)(x_i-\tilde{\mu}_c)^T}{n_c},\\
   \nonumber
\tilde{S}_c^{(U)}=&\frac{\sum_{i\in \Us}\alpha_{ic}(x_i-\tilde{\mu}_c)(x_i-\tilde{\mu}_c)^T}{\sum_{i\in \Us}\alpha_{ic}},\\
   \nonumber
\tilde{S}_c=&\frac{n_c\tilde{S}_c^{(L)}+(\sum_{i\in \Us}\alpha_{ic})\tilde{S}_c^{(U)}}{\tilde{n}_c}.
\end{align}
Then the profile of the negative penalized complete data log-likelihood for the $\Theta$'s replacing $\mu_c$ with $\tilde{\mu}_c$ and $\pi_c$ with $\tilde{\pi}_c$ is
\begin{equation}
\label{PEM}
\sum_{c\in \Cs}\tilde{n}_c\left\{\tr(\tilde{S}_c\Theta_c)-\log\det(\Theta_c)\right\}+\frac{\lambda_1}{j}\sum_{c \in \Cs}\left\vert\Theta_c\right\vert_j^j+\frac{\lambda}{j^2}\sum_{(c,m)\in \Cs \times \Cs}\left\vert\Theta_c-\Theta_m\right\vert_j^j,
\end{equation}
and maximizing this subject to $\Theta_c \in \Sp^p_+$ gives estimates of
the $\Theta$'s for the next iteration, estimates of the other parameters
for the next iteration being given by \eqref{eq:pi-twiddle}
and \eqref{eq:mu-twiddle}.
In the $j = 1$ case of \eqref{PEM} solutions are found by
the FGL algorithm \citep{JGL}, and in the $j=2$ case solutions are found by
our coordinate descent algorithm (Algorithm~\ref{thealg}).
In our current implementation, both algorithms are run until convergence.  

All of the steps above are repeated until the penalized EM (PEM) algorithm
converges.  Our convergence criterion here is similar to the one in
Section \ref{algsection}, in particular, we are using the difference in
the $\alpha$'s from iteration to iteration.
\citet{PEMcite} gives convergence rates for the penalized EM algorithm that
vary with the proportion of unlabeled data (the more unlabeled data the worse
the convergence).  Thus PEM should work well when the proportion of the
unlabeled data is not too large.  The initial estimates for our EM algorithm
are obtained from the labeled data \citep{Seeding}. 

An alternative is to not iterate to convergence in the optimization of
\eqref{PEM}.  \citet{DLR} call a variant of the EM algorithm in which
the M-step is not iterated to convergence but does make progress (goes uphill
on the function it is optimizing) a generalized EM (GEM) algorithm and 
\citet{EMAlg} proves this also converges to the MLE (under certain conditions).
The analog here, not iterating the M-step to convergence, is penalized
generalized EM (PGEM) and should also converge to the penalized maximum
likelihood estimate, although we have not investigated this.

\subsection{Validation Likelihood for Tuning Parameter Selection}
\label{sec:SSTune}
In the semi-supervised setting it is not uncommon to have data in which the labeled sample size for each class is so small that it would not be practical to use the validation likelihood presented in section \ref{secTune} to select the tuning parameters. To address this we propose a  validation likelihood that uses both labeled and unlabeled data.
The negative log-likelihood of the observed data $\Ds$ with parameters $\Psi$ is
\begin{equation}
\label{VLike}
L_{\Ds}\left(\Psi\right)=-\sum_{i \in \Ls} \log\{\pi_{y_i}\phi(x_i;\mu_{y_i},\Theta_{y_i})\}-\sum_{i \in \Us}\log\left\{\sum_{c \in \Cs} \pi_c\phi(x_i;\mu_c,\Theta_c)\right\}.
\end{equation}
Similar to the supervised case, randomly split the labeled and unlabeled data into $K$ subsets. We define $\Ls_{(v)}$ and $\Us_{(v)}$ to be the indices of the $v$th subset of the labeled and unlabeled data and $\Ds_{(v)}$ to be the $v$th subset of the data.   Let $\widehat{\Psi}_{(-v)}$ denote the parameter estimates resulting from the semi-supervised model based clustering on the data with
$\Ds_{(v)}$ removed.  

The validation likelihood is $L_{\Ds_{(v)}}(\widehat{\Psi}_{(-v)})$,
which is the negative log-likelihood for the $v$th subset of the data with
parameters estimates derived from all the data except  that subset
The validation score is
\begin{equation}
\label{SSVL}
V(\lambda_1,\lambda_2)=\sum_{v=1}^K L_{\Ds_{(v)}}\left(\widehat{\Psi}_{(-v)}\right)
\end{equation}
where $\widehat{\Psi}_{(-v)}$ are the parameter estimates based on $\lambda_1$ and $\lambda_2$ though the notation does not say this specifically.
We select the tuning parameters $\hat{\lambda}_1$ and $\hat{\lambda}_2$
that minimize \eqref{SSVL} over the set of allowed tuning parameter values.

\section{Simulations}
\label{SimStudy}
\subsection{Regularization in quadratic discriminant analysis}
\label{SupSim}
We present simulation studies that compare the classification performance of QDA in which
RDA, FGL, and the ridge fusion methods are used to estimate the inverse covariance matrices.  

\subsubsection{The data generating model and performance measurements}
\label{sec-datagen-qda}
In the following simulations described in sections \ref{SecSim1} -- \ref{SecSim6}, 
we generated data from a two-class model where the class 1 distribution was
$N_p(\mu_1, \Sigma_1)$ and the class 2 distribution was $N_p(\mu_2, \Sigma_2)$.  
We considered $p=50$ and $p=100$.
The training data had 25 independent draws from the class 1 distribution 
and 25 independent draws from the class 2 distribution.
These training 
observations were used to compute parameter estimates.  These estimates
were used in QDA to classify 
observations in an independent testing dataset consisting of 500 independent draws from 
the class 1 distribution and 500 independent draws from the class 2 distribution.
We measured performance with the classification error rate (CER) on these testing cases.
This process was replicated 100 times.

The tuning parameters $\lambda_1$ and $\lambda_2$ for FGL and the ridge fusion estimates of $\Sigma_{1}^{-1}$
and $\Sigma_{2}^{-1}$ were selected from a subset of  
$\{10^x: x=-10, -9.5,\ldots, 9, 9.5,10\}$ using the method described in 
section \ref{secTune} unless otherwise stated. Specific subsets were determined  
from pilot tests for each simulation. An R package, \verb@RidgeFusion@, implementing the ridge fusion and tuning parameter selection methods is available on CRAN \citep{RPackage}.

\subsubsection{RDA tuning parameter selection simulation}
\label{SecSim1}
In this simulation, we compared two cross-validation procedures to select
tuning parameters for the RDA estimators of $\Sigma_{1}^{-1}$ and $\Sigma_{2}^{-1}$.
The first procedure minimizes the validation CER and the second maximizes the validation likelihood, 
as described in section \ref{secTune}.
\citet{klaR}, in the documentation of the R package \verb@klaR@, mentioned that cross validation minimizing validation CER is unstable when sample sizes are small.  We used the klaR package to perform RDA with tuning parameter selection that
minimizes validation CER, and we used our own code to perform RDA with tuning parameter selection that maximizes validation likelihood.

We set all elements of $\mu_1$ to $5 p^{-1}\log(p)$ and made $\mu_2$ the vector of zeros. 
We generated $\Sigma_1$ and $\Sigma_2$ to have the same eigenvectors, which were the right singular vectors of the 100 by $p$ 
matrix with rows independently drawn from $N_p(0,I)$.  The $j$th eigenvalue of $\Sigma_1$ is 
$$
100\frac{p-j+1}{p} {\rm I}\{1\leq j \leq 6\} + 10\frac{p-j+1}{p} {\rm I}\{7\leq j \leq 11\}+ \frac{p-j+1}{p} {\rm I}\{12\leq j \leq p\}.
$$
The $j$th eigenvalue of $\Sigma_2$ is 
$$
500\frac{p-j+1}{p} {\rm I}\{1\leq j \leq 6\} + 50\frac{p-j+1}{p} {\rm I}\{7\leq j \leq 11\}+ \frac{p-j+1}{p} {\rm I}\{12\leq j \leq p\}.
$$
We investigated cases where $p=20,50,100$.  The results of this simulation, found in Table \ref{Sim1}, indicate that cross validation maximizing validation likelihood outperforms cross validation minimizing CER.  This lead us to tune RDA with the validation likelihood method in the remaining simulation studies.
\begin{table}
\caption{Average CER for RDA reported with standard errors based on 100 independent replications for the
simulation described in section \ref{SecSim1} (the RDA tuning parameter selection simulation). }
\label{Sim1}
\begin{center}
\begin{tabular}{l|ccc}
 & $p=20$ & $p=50$ & $p=100$ \\ 
\hline 
Validation Likelihood & 0.02 (0.01) & 0.05 (0.02) & 0.13 (0.04) \\  
Cross Validation with CER & 0.07 (0.03) & 0.09 (0.03) & 0.13 (0.04) \\ 
\end{tabular} 
\end{center}
\end{table}

\subsubsection{Dense, ill conditioned, and unequal inverse covariance matrices simulation: part 1}
\label{SecSim2}
This simulation uses the parameter values described in section \ref{SecSim1}
to  compare the QDA classification performance
of FGL, RDA, and the ridge fusion methods.  
Since $\Sigma_{1}^{-1}$ and $\Sigma_{2}^{-1}$
are dense, it is unclear which method should perform the best. 
Based on section \ref{ridgevsrda}, we expect that RDA will perform poorly 
because $\Sigma_1$ and $\Sigma_2$ are ill conditioned.
Table \ref{Sim2} has the average CER and corresponding standard errors. 
The ridge fusion method outperforms RDA.  We also see that ridge fusion and FGL perform similarly.

\begin{table}
\caption{Average CER for QDA with standard errors based on 100 independent replications for 
the simulation described in section \ref{SecSim2} (the dense, ill conditioned, and unequal inverse covariance matrices simulation: part 1).}
\label{Sim2}
\begin{center}
\begin{tabular}{l|cc} 
& $p=50$ & $p=100$ \\ 
\hline 
RDA & 0.05 (0.02)  & 0.13 (0.04)  \\ 
Ridge &0.03 (0.02)  & 0.08 (0.03) \\ 
FGL & 0.03 (0.02)  & 0.09  (0.02) \\ 
\end{tabular} 
\end{center}
\end{table}

\subsubsection{Dense, ill conditioned, and unequal inverse covariance matrices simulation:
part 2}
\label{SecSim6}
In this simulation, $\Sigma_1$ has $(i,j)$th entry $0.5 \cdot 1(|i-j|=1) + 1(i=j)$,
and $\Sigma_2$ is defined in section \ref{SecSim1}.
We set each element in $\mu_1$ to $p^{-1}$ and each element of $\mu_2$ to be zero.  
We expect RDA to perform poorly because of 
the large condition numbers and lack of similarity between $\Sigma_1$ and $\Sigma_2$.
The average classification error rate is reported in Table \ref{Sim6}, where we 
see that ridge fusion and FGL outperform RDA for both values of $p$.

\begin{table}
\caption{Average CER for QDA reported with standard errors based on 100 replications for the simulation described in section \ref{SecSim6} (the dense, ill conditioned, and unequal inverse covariance matrices simulation: part 2).}
\label{Sim6}
\begin{center}
\begin{tabular}{c|c|c}
• & $p=50$ & $p=100$ \\ 
\hline 
Ridge & 0.00 (0.00) & 0.00 (0.00) \\ 
RDA & 0.16 (0.04) & 0.31 (0.05) \\ 
FGL & 0.00 (0.00) & 0.00 (0.00) \\ 
\end{tabular} 
\end{center}
\end{table}

\subsubsection{Sparse, well conditioned, and equal inverse covariance matrices simulation}
\label{SecSim3}
In this simulation we set $\Sigma_1=\Sigma_2=I$ and all elements of $\mu_1$ to $10 p^{-1}\log(p)$ and all elements of $\mu_2$ to zero.
The average CER, based on 100 replications, is reported in Table \ref{Sim3}: 
all three methods perform similarly when $p=50$ and the ridge fusion method is outperformed 
by RDA and FGL when $p=100$.  

\begin{table}
\caption{Average CER for QDA reported with standard errors based on 100 independent replications for
the simulation described in section \ref{SecSim3} (the sparse, well conditioned, and equal inverse covariance matrices simulation).}
\label{Sim3}
\begin{center}
\begin{tabular}{l|cc}
& $p=50$ & $p=100$ \\ 
\hline 
RDA & 0.01 (0.01) & 0.03 (0.02) \\ 
Ridge & 0.01 (0.01) & 0.04  (0.02)\\  
FGL & 0.01 (0.01)& 0.03 (0.02)\\ 
\end{tabular} 
\end{center}
\end{table}

\subsubsection{Sparse and similar inverse covariance matrices simulation} 
\label{SecSim4}
In this simulation, $\Sigma_1$ 
is block diagonal with two equal size blocks: the $(i,j)$th entry in the first block
was $0.95^{|i-j|}$ and the $(k,m)$th entry in the second block was $0.8^{|k-m|}$.
We also made $\Sigma_2$ block diagonal 
with two equal size blocks: the $(i,j)$th entry in the first
block was $0.95^{|i-j|}$ and the $(k,m)$th entry in the second block was $\rho^{|k-m|}$, 
where $\rho = 0.25,\, 0.50,$ and $0.95$.  This setting should favor FGL, which exploits the sparsity
in $\Sigma_{1}^{-1}$ and $\Sigma_{2}^{-1}$.  
We set each element in 
$\mu_1$ to $20p^{-1}\log(p)$ and each element in $\mu_2$ to zero. 
The classification performance is reported in Table \ref{Sim4}. 
We see that FGL outperforms the other two methods for $\rho=0.25, 0.50$ and all values of $p$.  
When $\rho=0.95$, even though the covariance matrices are ill conditioned, 
RDA outperforms the ridge fusion method and  FGL for both values of $p$.

\begin{table}
\caption{Average CER for QDA reported with standard errors based on 100 independent replications for the simulation described in 
section \ref{SecSim4} (the sparse and similar inverse covariance matrices simulation).}
\label{Sim4}
\begin{center}
\begin{tabular}{l|c|cc}
 & $\rho$ & $p=50$ & $p=100$ \\ 
\hline 
RDA & & 0.10 (0.03) & 0.21 (0.04)  \\ 
 Ridge&0.95 & 0.13 (0.04) & 0.24 (0.04) \\ 
 FGL& & 0.11 (0.03) & 0.21 (0.04) \\ 
\hline 
 RDA & & 0.08 (0.03) & 0.20 (0.04) \\ 
 Ridge& 0.50 & 0.06 (0.02) & 0.13 (0.04)  \\ 
FGL& & 0.04 (0.02)  & 0.09 (0.03)\\ 
\hline 
 RDA& & 0.06 (0.02) & 0.15 (0.04) \\ 
 Ridge& 0.25 & 0.05 (0.02) & 0.12 (0.03) \\ 
 FGL& & 0.03 (0.02) & 0.06 (0.02) \\ 
\end{tabular} 
\end{center}
\end{table}

\subsubsection{Inverse covariance matrices with small entries simulation}
\label{SecSim5}
In this simulation, $\Sigma_1$ has $(i,j)$th entry $0.4 \cdot 1(|i-j|=1) + 1(i=j)$
and $\Sigma_2$ has $(i,j)$th entry $\rho \cdot 1(|i-j|=1) + 1(i=j)$, where
$\rho=0.25, 0.30, 0.35$ and $0.50$. 
We set each element in $\mu_1$ to $10\log(p)p^{-1}$ and each element in $\mu_2$ to zero. 
The classification results are reported in Table \ref{Sim5}, 
and show that RDA has the best classification performance for each value of $p$ and $\rho$.  
Note that FGL has the same average classification error rate as RDA in the case where $p=50$ when $\rho=0.30$ and $0.35$.

\begin{table}
\caption{Average CER for QDA reported with standard errors based on 100 replications
 for the simulation described in section \ref{SecSim5} (the inverse covariance matrices with small entries simulation).}
\label{Sim5}
\begin{center}
\begin{tabular}{c|c|c|c}
• & $\rho$ & $p=50$ & $p=100$ \\ 
\hline 
Ridge & • & 0.04 (0.02) & 0.09 (0.03)  \\ 
RDA & 0.25 & 0.02 (0.01)  & 0.06 (0.02) \\ 
FGL & • & 0.03 (0.02) & 0.07 (0.03) \\ 
\hline 
Ridge & • & 0.04 (0.02)  & 0.09 (0.03)  \\ 
RDA & 0.30 & 0.03 (0.02)  & 0.06 (0.02)  \\ 
FGL &  &0.03 (0.02)  & 0.08 (0.03)\\ 
\hline 
Ridge & • & 0.04 (0.02)   & 0.10 (0.03)  \\ 
RDA & 0.35 & 0.03 (0.02)  & 0.07 (0.03)  \\ 
FGL & • & 0.03 (0.02)  & 0.08 (0.03) \\ 
\hline 
Ridge& &0.06 (0.02) & 0.11 (0.03)\\
RDA& 0.50& 0.03 (0.02) & 0.09 (0.03)\\
FGL& &0.04 (0.02) &0.10 (0.03)\\
\end{tabular} 
\end{center}
\end{table}

\subsection{Computing time simulations: ridge fusion versus FGL}
\label{Timing Sims}
Although FGL performed as well or better than our ridge fusion method at classification
in the simulations of sections \ref{SecSim2} -- \ref{SecSim6}, we found that computing
FGL is much slower than our ridge fusion method when a dense estimate is desired.  
We present three timing simulations that illustrate this pattern.
In each simulation we measured the computing time (in seconds) of ridge fusion and FGL, calculated by the R function \verb@system.time@, where the tuning parameters $(\lambda_1,\lambda_2)$ 
are selected from  $\Lambda \times \Lambda$, 
where $\Lambda = \{10^x: x=-8,-7,\ldots, 7, 8\}$ and $p=100$. We report the average of the difference in  computing time between ridge fusion and FGL based on 100 independent replications, for each point in $\Lambda\times\Lambda$.
FGL and ridge fusion were computed using the
\verb@JGL@ \citep{JGLR} and \verb@RidgeFusion@ \citep{RPackage} R packages
with default settings.

In each simulation setting,  the ridge fusion algorithm is faster than FGL when $\lambda_1$ is small, and FGL is faster than ridge fusion when $\lambda_1$ is large.  This result is not surprising as a large $\lambda_1$  when using FGL will produce sparse estimates of the inverse covariance matrices, which the algorithm exploits in estimation by using a divide and conquer algorithm.
In summary, FGL will be faster when the true inverse covariance matrices are
quite sparse and otherwise ridge fusion will be faster.

\subsubsection{Dense, ill conditioned, and different inverse covariance matrices timing simulation}
\label{SecTiming1}
\begin{figure}
\begin{center}
\includegraphics[width=3in]{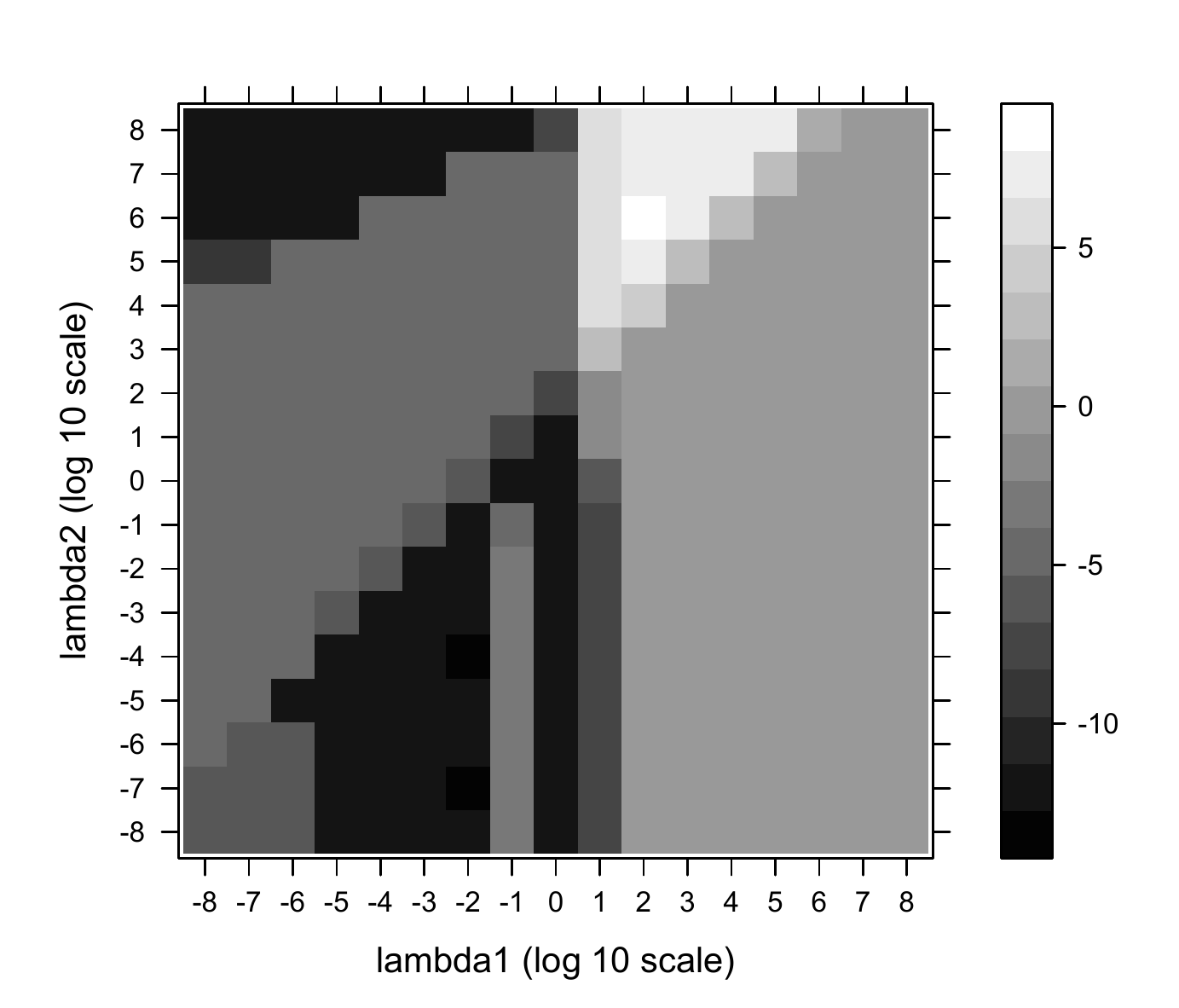}
\end{center}
\caption{Difference of average computing time for the simulation described in section \ref{SecTiming1} based on 100 replications at each point in $\Lambda\times\Lambda$.  
Negative values represent where ridge fusion is faster than FGL.}
\label{TimingFig1}
\end{figure}
This simulation investigates the difference of the average speed over 100 replications of FGL and ridge fusion using the data generating model described in section \ref{sec-datagen-qda} and parameter values 
used in section \ref{SecSim2}.  The results are shown in in Figure \ref{TimingFig1}. The ridge fusion method is faster or comparable to FGL when $\lambda_1$ is small and otherwise FGL is faster.  Over the entire grid we find that, on average, ridge fusion is 4 seconds faster than FGL.  At one grid point, ridge fusion was $534$ times faster than FGL and at another grid point FGL was $73$ times faster than ridge fusion.

\subsubsection{Sparse and similar inverse covariance matrices timing simulation}
\label{SecTiming2}
\begin{figure}
\begin{center}
\includegraphics[width=3in]{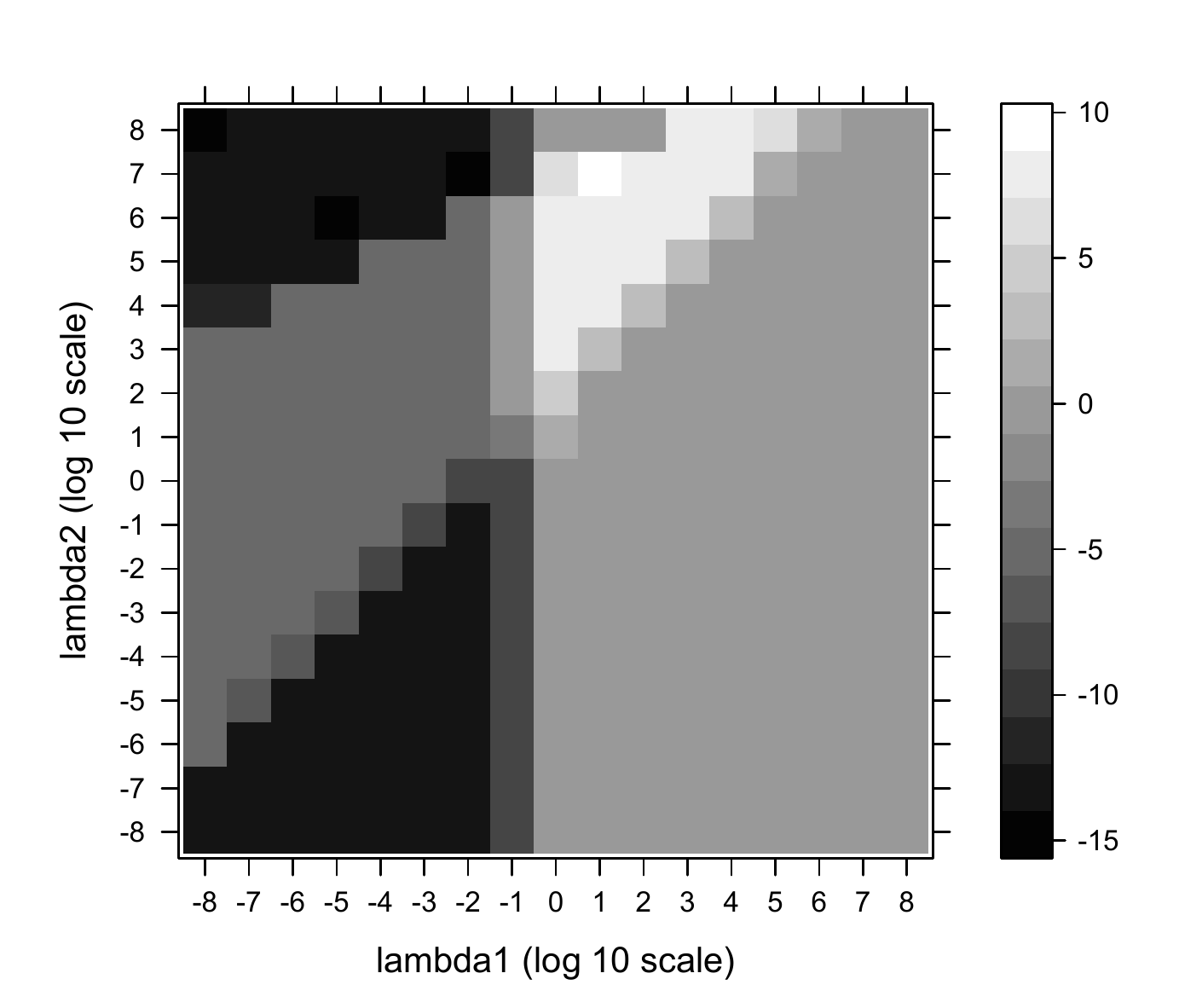}
\end{center}
\caption{Difference of average computing time for the simulation described in section \ref{SecTiming2} based on 100 replications at  each point in $\Lambda\times\Lambda$. 
Negative values represent where ridge fusion is faster than FGL.}
\label{TimingFig2}
\end{figure}
This simulation uses the data generating model described in section \ref{sec-datagen-qda} and parameter values 
used in section \ref{SecSim4} with $\rho=0.95$ and $p=100$.  Here FGL performs much better than ridge fusion in classification.  The results shown in Figure \ref{TimingFig2} are the difference of the average computing time of FGL and ridge fusion.  These are similar to those of section \ref{SecTiming1}.
As expected, ridge fusion is faster or comparable to FGL when $\lambda_1$ is small and otherwise FGL is faster.  Averaging across the grid, ridge fusion was approximately 5 seconds faster.  At the extremes,
there was one grid point at which ridge fusion was $564$ times faster than FGL, and  another point at which FGL was $63$ times faster than ridge fusion.

\subsubsection{Inverse covariance matrices with small entries timing simulation}
\label{timing-dense}
\begin{figure}
\begin{center}
\includegraphics[width=3in]{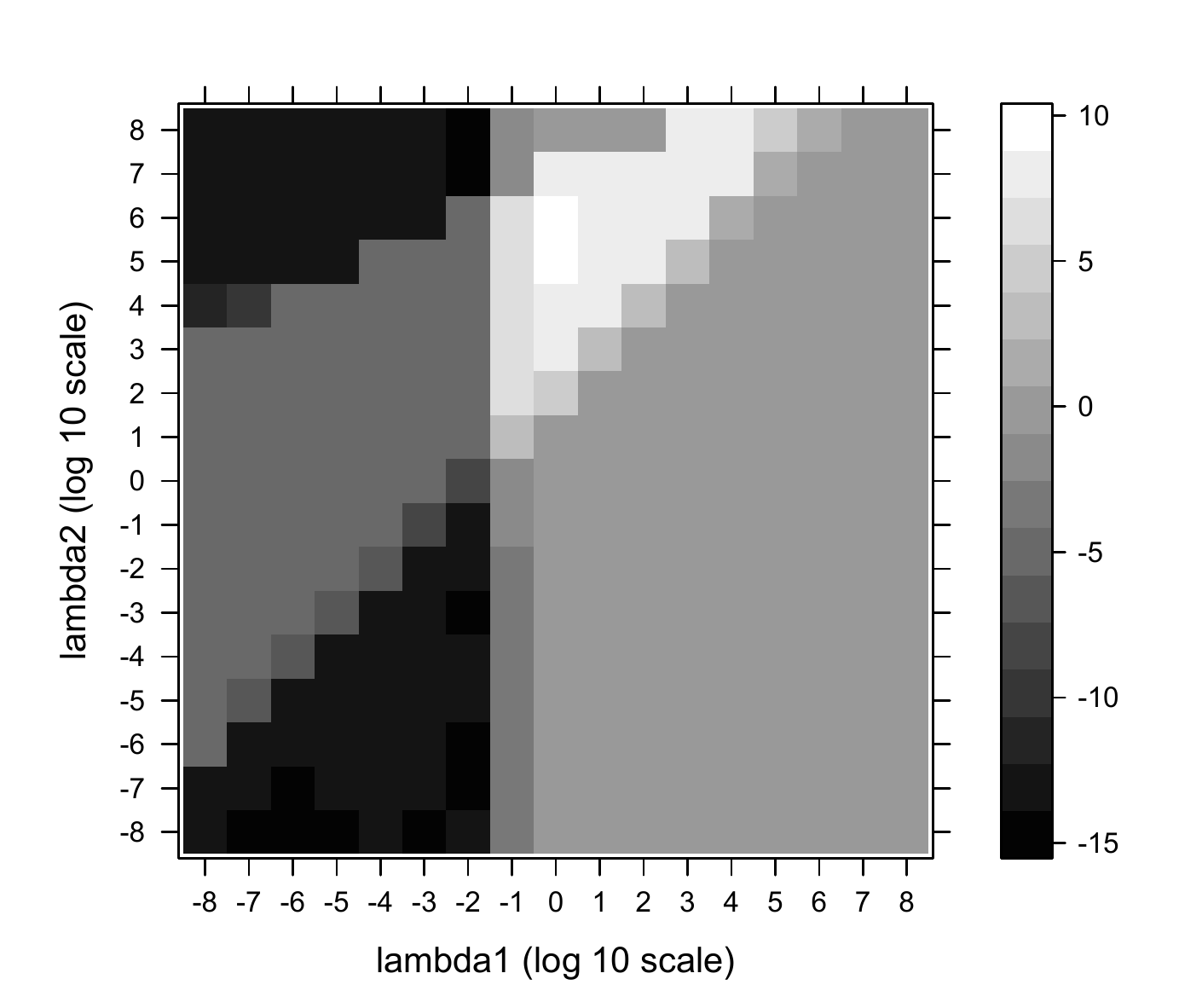}
\end{center}
\caption{Difference of average computing time for the simulation described in section \ref{timing-dense} based on 100 replications at each point in $\Lambda\times\Lambda$.  
Negative values represent where ridge fusion is faster than FGL.}
\label{TimingFig3}
\end{figure}

This simulation uses the data generating model described in section \ref{sec-datagen-qda} and parameter values 
used in section \ref{SecSim5} when $\rho=0.50$ and $p=100$.  The results in Figure \ref{TimingFig3} 
show a similar result to the other timing simulations in sections \ref{SecTiming1} and \ref{SecTiming2}:
ridge fusion is faster or comparable to FGL when $\lambda_1$ is small and otherwise FGL is faster.  We find that on average over the entire grid that ridge fusion is on average 5 seconds faster.  At the extremes, 
there was one point on the grid where ridge fusion was $595$ times faster than FGL and
another point on the grid where FGL was $322$ times faster than ridge fusion.

\subsection{Regularization in semi-supervised model based clustering}
\label{SemiSupSim}
We evaluate the semi-supervised model based clustering methods proposed in section \ref{sec:SSMBC} by 
comparing the tuning parameter selection methods proposed in
sections~\ref{sec:SSTune} and~\ref{secTune}.
This simulation uses the same data generating model as that used in the simulation study in section \ref{SecSim4} with $\rho=0.25$. 
Each replication will have 25 labeled and 250 unlabeled observations from each class.  We compare the ridge fusion and FGL methods for semi-supervised model based clustering on their ability to classify the unlabeled data for 50 independent replications when the tuning parameters are selected using the labeled data only via the methodology proposed in section \ref{sec:SSTune}.  For each replication the QDA classification rule is formed by using the corresponding parameter estimates from the regularized semi-supervised model based clustering.  Results of this simulation are contained in Table \ref{SimSS} and show that using the method presented in section \ref{sec:SSTune} to select the tuning parameter outperforms the method
that ignores the unlabeled data.

\begin{table}
\caption{Average CER reported with standard errors for the semi-supervised model based clustering simulation based on 50 independent replications.}
\label{SimSS}
\begin{center}
\begin{tabular}{l|cc}
 & $p=50$ & $p=100$ \\ 
\hline 
Ridge &0.01 (0.01) & 0.07 (0.04)  \\ 
Ridge Labeled & 0.02 (0.02) & 0.22 (0.06)\\ 
FGL &  0.01 (0.01) & 0.01 (0.01) \\
FGL Labeled&0.04 (0.03) & 0.31 (0.07)
\end{tabular} 
\end{center}
\end{table}

\section{Data Example}
We compare ridge fusion, FGL and RDA on the Libras movement data from the UCI Machine Learning Repository, 
which describes the  hand movements in Brazilian sign language \citep{UCIRep}.  The original data has 15 classes corresponding to the type of hand movements, with 90 variables that represent 45 different time points in a video that shows the hand movement.  The variables represent where the hand is in the frame at a given time point.  For this example we selected 3 classes that correspond to the movements of curved swing, horizontal swing, and vertical swing. We have taken 18 observations from each class for training while keeping 45 observations from each class for validation.  Tuning parameters for each method were selected using 3-fold validation likelihood due to the small sample size of the training set.  The results of this analysis are reported in Table \ref{LIBRAStable}, and show that the ridge fusion method outperforms FGL and RDA with regard to the number of observations classified incorrectly for the validation data.

\begin{table}
\caption{Fraction of the validation data that is classified incorrectly for the Libra data example.}
\label{LIBRAStable}
\begin{center}
\begin{tabular}{l|c}
& Fraction of Data Misclassified \\ 
\hline 
Ridge & 0/135 \\  
FGL & 9/135 \\ 
RDA & 4/135 \\  
\end{tabular} 
\end{center}
\end{table}

We also apply the methodology from section \ref{sec:SSMBC} on the Libras movement data where the 45 validation points from each class are treated as the unlabeled data.  Again we use a 3-fold validation likelihood based on the method proposed in section \ref{sec:SSTune}.  Table \ref{SSLibras} contains the results. As we saw in the supervised case, the ridge fusion method has a smaller number of observations classified incorrectly when compared to FGL on the unlabeled data.

\begin{table}
\caption{Fraction of the unlabeled data that is classified incorrectly using semi-supervised model based clustering methods for the Libra data example.}
\label{SSLibras}
\begin{center}
\begin{tabular}{l|c}
& Fraction of Unlabeled Data Misclassified\\
\hline
Ridge& 0/135\\
FGL& 5/135\\
\end{tabular}
\end{center}
\end{table}

\section{Acknowledgments}
This research is partially supported by the National Science Foundation grant
DMS-1105650.

\bibliographystyle{jcgs}
\bibliography{JGL}

\begin{thebibliography}{18}
\newcommand{\enquote}[1]{``#1''}
\providecommand{\natexlab}[1]{#1}
\providecommand{\url}[1]{\texttt{#1}}
\providecommand{\urlprefix}{URL }

\bibitem[{Bache and Lichman(2013)}]{UCIRep}
Bache, K. and Lichman, M. (2013), \enquote{{UCI} Machine Learning Repository,}
  \url{http://archive.ics.uci.edu/ml}.

\bibitem[{Basu et~al.(2002)Basu, Banerjee, and Mooney}]{Seeding}
Basu, S., Banerjee, A., and Mooney, R. (2002), \enquote{Semi-supervised
  Clustering by Seeding,} in \emph{Proceedings of the 19th International
  Conference on Machine Learning}, 19--26.

\bibitem[{Danaher(2013)}]{JGLR}
Danaher, P. (2013), \emph{JGL: Performs the Joint Graphical Lasso for sparse
  inverse covariance estimation on multiple classes},
  \urlprefix\url{http://CRAN.R-project.org/package=JGL}, r package version 2.3.

\bibitem[{Danaher et~al.(2013)Danaher, Wang, and Witten}]{JGL}
Danaher, P., Wang, P., and Witten, D. (2013), \enquote{The Joint Graphical
  Lasso for Inverse Covariance Estimation Across Multiple Classes,} \emph{The
  Journal of Royal Statistical Society, Series B}.

\bibitem[{Dempster et~al.(1977)Dempster, Laird, and Rubin}]{DLR}
Dempster, A., Laird, N., and Rubin, D. (1977), \enquote{Maximum Likelihood From
  Incomplete Data via the EM Algorithm (with discussion),} \emph{Journal of The
  Royal Statistical Society, Series B}, 39, 1--38.

\bibitem[{Friedman(1989)}]{RegDiscrim}
Friedman, J. (1989), \enquote{Regularized Discriminant Analysis,} \emph{Journal
  of the American Statistical Association}, 84, 249--266.

\bibitem[{Green(1990)}]{PEMcite}
Green, P. (1990), \enquote{On Use of the EM for Penalized Likelihood
  Estimation,} \emph{Journal of the Royal Statistical Society, Series B}, 52,
  443--452.

\bibitem[{Guo et~al.(2011)Guo, Levina, Michailidis, and
  Zhu}]{JointEstimationMGM}
Guo, J., Levina, E., Michailidis, G., and Zhu, J. (2011), \enquote{Joint
  Estimation of Multiple Graphical Models,} \emph{Biometrika}, 98, 1--15.

\bibitem[{Huang et~al.(2006)Huang, Liu, Pourahmadi, and Liu}]{huang06}
Huang, J., Liu, N., Pourahmadi, M., and Liu, L. (2006), \enquote{Covariance
  Matrix Selection and Estimation via Penalised Normal Likelihood,}
  \emph{Biometrika}, 93, 85--98.

\bibitem[{Pourahmadi(2011)}]{Pour11}
Pourahmadi, M. (2011), \enquote{Covariance Estimation: The GLM and
  Regularization Perspective,} \emph{Statistical Science}, 26, 369--387.

\bibitem[{Price(2014)}]{RPackage}
Price, B.~S. (2014), \emph{RidgeFusion: R Package for Ridge Fusion in
  Statistical Learning}, {R} package version 1.0-3.

\bibitem[{Rothman and Forzani(2013)}]{RothmanForzani13}
Rothman, A.~J. and Forzani, L. (2013), \enquote{Properties of optimizations
  used in penalized Gaussian likelihood inverse covariance matrix estimation,}
  Manuscript.

\bibitem[{Ruan et~al.(2011)Ruan, Yuan, and Zou}]{L1MBC}
Ruan, L., Yuan, M., and Zou, H. (2011), \enquote{Regularized Parameter
  Estimation in High-Dimensional Gaussian Mixture Models,} \emph{Neural
  Computation}, 23, 1605--1622.

\bibitem[{Weihs et~al.(2005)Weihs, Ligges, Luebke, and Raabe}]{klaR}
Weihs, C., Ligges, U., Luebke, K., and Raabe, N. (2005), \enquote{klaR
  Analyzing German Business Cycles,} in \emph{Data Analysis and Decision
  Support}, eds. D.~Baier, R.~Decker, and L.~Schmidt-Thieme, Berlin:
  Springer-Verlag, 335--343.

\bibitem[{Witten and Tibshirani(2009)}]{Scout}
Witten, D. and Tibshirani, R. (2009), \enquote{Covariance Regularized
  Regression and Classification for High-Dimensional Problems,} \emph{Journal
  of Royal Statistical Society, Series B}, 71, 615--636.

\bibitem[{Wu(1983)}]{EMAlg}
Wu, C.~J. (1983), \enquote{On the Convergence Properties of the EM Algorithm,}
  \emph{Annals of Statistics}, 11, 95--103.

\bibitem[{Xie et~al.(2008)Xie, Pan, and Shen}]{DiagCluster}
Xie, B., Pan, W., and Shen, X. (2008), \enquote{Penalized Model Based
  Clustering with Cluster Specific Diagonal Covariance Matrices and Grouped
  Variables,} \emph{Electronic Journal of Statistics}, 2, 168--212.

\bibitem[{Zhou et~al.(2009)Zhou, Pan, and Shen}]{UnconMBC}
Zhou, H., Pan, W., and Shen, X. (2009), \enquote{Penalized Model-Based
  Clustering with Unconstrained Covariance Matrices,} \emph{Electronic Journal
  of Statistics}, 3, 1473--1496.

\end{thebibliography}

\end{document}